\def\year{2019}\relax
\documentclass[letterpaper]{article} 
\usepackage{aaai19}  
\usepackage{times}  
\usepackage{helvet}  
\usepackage{courier}  
\usepackage{url}  
\usepackage{graphicx}  
\usepackage{booktabs}
\usepackage{algorithm}
\usepackage{algorithmic}
\usepackage{multirow}
\usepackage{amsmath,amscd,amsbsy,amssymb,latexsym,url,bm}

\DeclareMathOperator*{\argmax}{argmax}
\newcommand{\tabincell}[2]{\begin{tabular}{@{}#1@{}}#2\end{tabular}}
\newcommand{\citet}[1]{\citeauthor{#1}~\shortcite{#1}}

\begin{document}

\title{Large-scale Interactive Recommendation \\ with Tree-structured Policy Gradient}

\author{Haokun Chen,\textsuperscript{1} Xinyi Dai,\textsuperscript{1} Han Cai,\textsuperscript{1} Weinan Zhang,\textsuperscript{1} \\
{\bf \Large Xuejian Wang,\textsuperscript{1} Ruiming Tang,\textsuperscript{2} Yuzhou Zhang,\textsuperscript{2} Yong Yu\textsuperscript{1}} \\
\textsuperscript{1}Shanghai Jiao Tong University \\
\textsuperscript{2}Huawei Noah's Ark Lab \\
\{chenhaokun,xydai,hcai,wnzhang,xjwang,yyu\}@apex.sjtu.edu.cn \\
\{tangruiming,zhangyuzhou3\}@huawei.com}
\maketitle

\begin{abstract}
	Reinforcement learning (RL) has recently been introduced to interactive recommender systems (IRS) because of its nature of learning from dynamic interactions and planning for long-run performance.
	As IRS is always with thousands of items to recommend (i.e., thousands of actions), most existing RL-based methods, however, fail to handle such a large discrete action space problem and thus become inefficient.
	The existing work that tries to deal with the large discrete action space problem by utilizing the deep deterministic policy gradient framework suffers from the inconsistency between the continuous action representation (the output of the actor network) and the real discrete action.
	To avoid such inconsistency and achieve high efficiency and recommendation effectiveness, in this paper, we propose a Tree-structured Policy Gradient Recommendation (TPGR) framework, where a balanced hierarchical clustering tree is built over the items and picking an item is formulated as seeking a path from the root to a certain leaf of the tree.
	Extensive experiments on carefully-designed environments based on two real-world datasets demonstrate that our model provides superior recommendation performance and significant efficiency improvement over state-of-the-art methods.
\end{abstract}

\section{Introduction}
Interactive recommender systems (IRS) \cite{zhao2013interactive} play a key role in most personalized services, such as Pandora, Musical.ly and YouTube, etc.
Different from the conventional recommendation settings \cite{mooney2000content,koren2009matrix}, where the recommendation process is regarded as a static one, an IRS consecutively recommends items to individual users and receives their feedbacks which makes it possible to refine its recommendation policy during such interactive processes.

To handle the interactive nature, some efforts have been made by modeling the recommendation process as a multi-armed bandit (MAB) problem  \cite{li2010contextual,zhao2013interactive}. However, these works pre-assume that the underlying user preference remains unchanged during the recommendation process \cite{zhao2013interactive} and do not plan for long-run performance explicitly.

Recently, reinforcement learning (RL) \cite{sutton1998reinforcement}, which has achieved remarkable success in various challenging scenarios that require both dynamic interaction and long-run planning such as playing games \cite{mnih2015human,silver2016mastering} and regulating ad bidding \cite{cai2017real,jin2018real}, has been introduced to model the recommendation process and shows its potential to handle the interactive nature in IRS \cite{zheng2018drn,zhao2018recommendations,zhao2018deep}.

However, most existing RL techniques cannot handle the large discrete action space problem in IRS as the time complexity of making a decision is linear to the size of the action space.
Specifically, all Deep Q-Network (DQN) based methods \cite{zheng2018drn,zhao2018recommendations} involve a maximization operation taken over the action space to make a decision, which becomes intractable when the size of the action space, i.e., the number of available items, is large \cite{dulac2015deep}, which is very common in IRS.
Most Deep Deterministic Policy Gradient (DDPG) based methods \cite{zhao2018deep,Hu2018Reinforcement} also suffer from the same problem as a specific ranking function is applied over all items to pick the one with highest score when making a decision.
To reduce the time complexity, \citet{dulac2015deep} propose to select the proto-action in a continuous hidden space and then pick the valid item via a nearest-neighbor method. 
However, such a method suffers from the inconsistency between the learned continuous action and the actually desired discrete action, and thereby may lead to unsatisfied results \cite{tavakoli2017action}.

In this paper, we propose a Tree-structured Policy Gradient Recommendation (TPGR) framework which achieves high efficiency and high effectiveness at the same time.
In the TPGR framework, a balanced hierarchical clustering tree is built over the items and picking an item is thus formulated as seeking a path from the root to a certain leaf of the tree, which dramatically reduces the time complexity in both the training and the decision making stages.
We utilize policy gradient technique \cite{sutton2000policy} to learn how to make recommendation decisions so as to maximize long-run rewards.
To the best of our knowledge, this is the first work of building tree-structured stochastic policy for large-scale interactive recommendation.

Furthermore, to justify the proposed method using public available offline datasets, we construct an environment simulator to mimic online environments with principles derived from real-world data. 
Extensive experiments on two real-world datasets with different settings show superior performance and significant efficiency improvement of the proposed TPGR over state-of-the-art methods.

\section{Related Work and Background}\label{sec:related}
\subsection{Advanced Recommendation Algorithms for IRS}
\subsubsection{MAB-based Recommendation}
A group of works \cite{li2010contextual,chapelle2011empirical,zhao2013interactive,zeng2016online,wang2016learning} try to model the interactive recommendation as a MAB problem.
\citet{li2010contextual} adopt a linear model to estimate the Upper Confidence Bound (UCB) for each arm.
\citet{chapelle2011empirical} utilize the Thompson sampling technique to address the trade-off between exploration and exploitation.
Besides, some researchers try to combine MAB with matrix factorization technique \cite{zhao2013interactive,kawale2015efficient,wang2017factorization}.

\subsubsection{RL-based Recommendation}
RL-based recommendation methods \cite{tan2017neural,zheng2018drn,zhao2018recommendations,zhao2018deep}, which formulate the recommendation procedure as a Markov Decision Process (MDP), explicitly model the dynamic user status and plan for long-run performance.
\citet{zhao2018recommendations} incorporate negative as well as positive feedbacks into a DQN framework \cite{mnih2015human} and propose to maximize the difference of Q-values between the target and the competitor items.
\citet{zheng2018drn} combine DQN and Dueling Bandit Gradient Decent (DBGD) \cite{grotov2016online} to conduct online news recommendation.
\citet{zhao2018deep} propose to utilize a DDPG framework \cite{lillicrap2015continuous} with a page-display approach for page-wise recommendation.

\subsection{Large Discrete Action Space Problem in RL-based Recommendation}
Most RL-based models become unacceptably inefficient for IRS with large discrete action space as the time complexity of making a decision is linear to the size of the action space.

For all DQN-based solutions \cite{zhao2018recommendations,zheng2018drn}, a value function $Q(s, a)$, which estimates the expected discounted cumulative reward when taking the action $a$ at the state $s$, is learned and the policy's decision is: 
\begin{equation}\label{eq:dqn_issue}
\pi_Q(s) = \argmax_{a \in A} Q(s, a).
\end{equation}

As shown in Eq.~(\ref{eq:dqn_issue}), to make a decision, $|A|$ ($A$ denotes the item set) evaluations are required, which makes both learning and utilization intractable for tasks where the size of the action space is large, which is common for IRS. 

Similar problem exists in most DDPG-based solutions \cite{zhao2018deep,Hu2018Reinforcement} where some ranking parameters are learned and a specific ranking function is applied over all items to pick the one with highest ranking score.
Thus, the complexity of sampling an action for these methods also grows linearly with respect to $|A|$. 

\citet{dulac2015deep} attempt to address the large discrete action space problem based on the DDPG framework by mapping each discrete action to a low-dimensional continuous vector in a hidden space while maintaining an actor network to generate a continuous vector $a_v$ in the hidden space which is later mapped to a specific valid action $a$ among the $k$-nearest neighbors of $a_v$. 
Meanwhile, a value network $Q(s, a)$ is learned using transitions collected by executing the valid action $a$ and the actor network is updated according to $\frac{\partial Q(s, \hat{a})}{\partial \hat{a}} \big|_{\hat{a} = a_v}$ following the DDPG framework.
Though such a method can reduce the time complexity of making a decision from $\mathcal{O}(|A|)$ to $\mathcal{O}(\log(|A|))$ when the value of $k$ (i.e., the number of nearest neighbors to find) is small, there is no guarantee that the actor network is learned in a correct direction as in the original DDPG.
The reason is that the value network $Q(s,a)$ may behave differently on the output of the actor network $a_v$ (when training the actor network) and the actually executed action $a$ (when training the value network). 
Besides, the utilized approximate $k$-nearest neighbors (KNN) method may also cause trouble as the found neighbors may not be exactly the nearest ones.

In this paper, we propose a novel solution to address the large discrete action space problem. Instead of using the continuous hidden space, we build a balanced tree to represent the discrete action space where each leaf node corresponds to an action and top-down decisions are made from the root to a specific leaf node to take an action, which reduces the time complexity of making a decision from $\mathcal{O}(|A|)$ to $\mathcal{O} (d \times |A|^{1/d})$, where $d$ denotes the depth of the tree. Since such a method does not involve a mapping from the continuous space to the discrete space, it avoids the gap between the continuous vector given by the actor network and the actually executed discrete action in \cite{dulac2015deep}, which could lead to incorrect updates. 

\section{Proposed Model}\label{sec:method}
\subsection{Problem Definition}\label{sec:prob_def}
We use an MDP to model the recommendation process, where the key components are defined as follows.
\begin{itemize}
	\item \textbf{State.} A state $s$ is defined as the historical interactions between a user and the recommender system, which can be encoded as a low-dimensional vector via a recurrent neural network (RNN) (see Figure~\ref{fig:state representation}).
	\item \textbf{Action.} An action $a$ is to pick an item for recommendation, such as a song or a video, etc.
	\item \textbf{Reward.} In our formulation, all users interacting with the recommender system form the environment that returns a reward $r$ after receiving an action $a$ at the state $s$, which reflects the user's feedback to the recommended item. 
	\item \textbf{Transition.} As the state is the historical interactions, once a new item is recommended and the corresponding user's feedback is given, the state transition is determined. 
\end{itemize}

An episode in the above defined MDP corresponds to one recommendation process, which is a sequence of user states, recommendation actions and user's feedbacks, e.g., $(s_1, a_1, r_1, s_2, a_2, r_2, \cdots,$ $s_n,$ $a_n, r_n, s_{n + 1})$. 
In this case, the sequence starts with user state $s_1$ and then transits to $s_2$ after a recommendation action $a_1$ is carried out by the recommender system and a reward $r_1$ is given by the environment indicating the user's feedback to the recommendation action.
The sequence is terminated at a specific state $s_{n+1}$ when some pre-defined conditions are satisfied. Without loss of generality, we set the length of an episode $n$ to a fixed number \cite{cai2017real,zhao2018recommendations}. 

\subsection{Tree-structured Policy Gradient Recommendation}
\subsubsection{Intuition for TPGR}
To handle the large discrete action space problem and achieve high recommendation effectiveness, we propose to build up a balanced hierarchical clustering tree over items (Figure~\ref{fig:TPGR} left) and then utilize the policy gradient technique to learn the strategy of choosing the optimal subclass at each non-leaf node of the constructed tree (Figure~\ref{fig:TPGR} right).
Specifically, in the clustering tree, each leaf node is mapped to a certain item (Figure~\ref{fig:TPGR} left) and each non-leaf node is associated with a policy network (note that only three but not all policy networks are shown in the right part of Figure~\ref{fig:TPGR} for the ease of presentation). As such, given a state and guided by the policy networks, a top-down moving is performed from the root to a leaf node and the corresponding item is recommended to the user.

\subsubsection{Balanced Hierarchical Clustering over Items}\label{sec:bhc}
Hierarchical clustering seeks to build a hierarchy of clusters, i.e., a clustering tree. 
One popular method is the divisive approach where the original data points are divided into several clusters, 
and each cluster is further divided into smaller sub-clusters.
The division is repeated until each sub-cluster is associated with only one point.

In this paper, we aim to conduct \emph{balanced} hierarchical clustering over items, where the constructed clustering tree is supposed to be balanced, i.e., for each node, the heights of its subtrees differ by at most one and the subtrees are also balanced. For the ease of presentation and implementation, it is also required that each non-leaf node has the same number of child nodes, denoted as $c$, except for parents of leaf nodes, whose numbers of child nodes are at most $c$.

We can perform balanced hierarchical clustering over items following a clustering algorithm which takes a group of vectors and an integer $c$ as input and divides the vectors into $c$ balanced clusters (i.e., the item number of each cluster differs from each other by at most one). In this paper, we consider two kinds of clustering algorithms, i.e., PCA-based and K-means-based clustering algorithms whose detailed procedures are provided in the appendices. 
By repeatedly applying the clustering algorithm until each sub-cluster is associated with only one item, a balanced clustering tree is constructed. As such, denoting the item set and the depth of the balanced clustering tree as $A$ and $d$ respectively, we have:
\begin{equation}\label{eq:node_relation}
c^{d - 1} < |A| \leq c^{d}.
\end{equation}

Thus, given $A$ and $d$, we can set $c = ceil(|A|^{\frac{1}{d}})$ where $ceil(x)$ returns the smallest integer which is no less than $x$.

The balanced hierarchical clustering over items is normally performed on the (vector) representation of the items, which may largely affect the quality of the attained balanced clustering tree. In this work we consider three approaches for producing such representation: 
\begin{itemize}
	\item \textbf{Rating-based.} An item is represented as the corresponding column of the user-item rating matrix, where the value of each element $(i, j)$ is the rating of user $i$ to item $j$. 
	\item \textbf{VAE-based.} Low-dimensional representation of the rating vector for each item can be learned by utilizing a variational auto-encoder (VAE) \cite{kingma2013auto}.
	\item \textbf{MF-based.} The matrix factorization (MF) technique \cite{koren2009matrix} can also be utilized to learn a representation vector for each item.  
\end{itemize}

\subsubsection{Architecture of TPGR}
\begin{figure}[!t]
	\centering
	\includegraphics[width=\columnwidth]{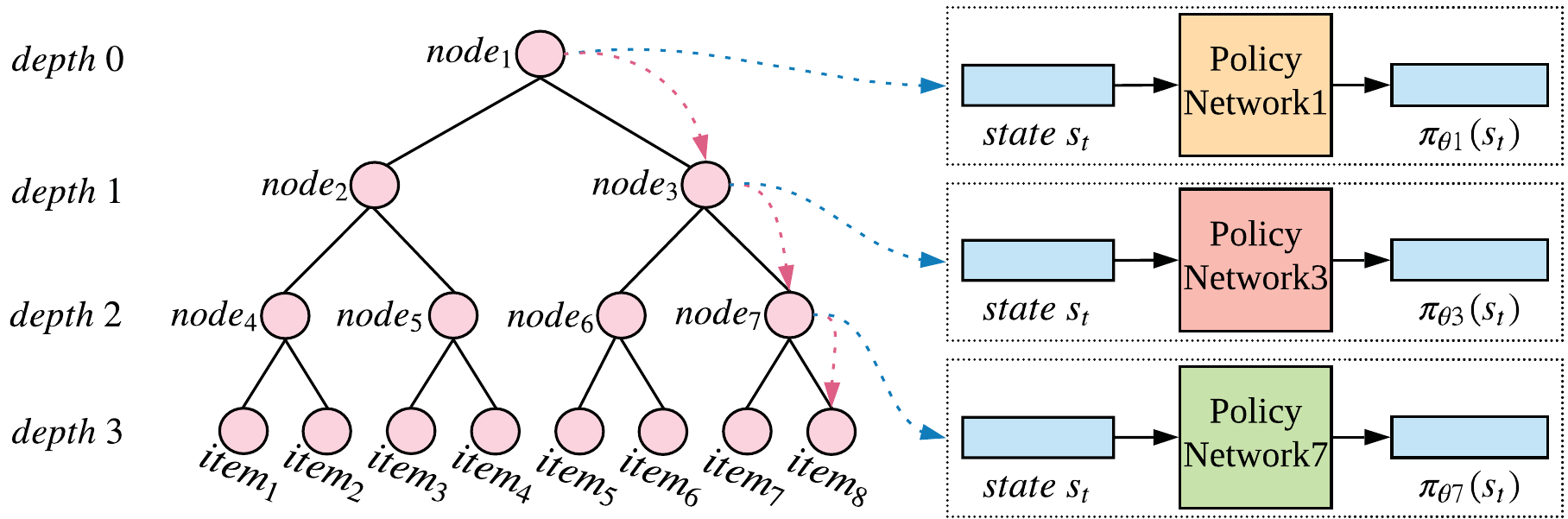}	\caption{Architecture of TPGR.}
	\label{fig:TPGR}
\end{figure}

The architecture of the Tree-structured Policy Gradient Recommendation (TPGR) is based on the constructed clustering tree.
To ease the illustration, we assume that there is a status point to indicate which node is currently located.
Thus, picking an item is to move the status point from the root to a certain leaf.
Each non-leaf node of the tree is associated with a policy network which is implemented as a fully-connected neural network with a softmax activation function on the output layer.
Considering node $v$ where the status point is located, the policy network associated with $v$ takes the current state as input and outputs a probability distribution over all child nodes of $v$, which indicates the probability of moving to each child node of $v$.

Using a recommendation scenario with 8 items for illustration, the constructed balanced clustering tree with the tree depth set to 3 is shown in Figure \ref{fig:TPGR} (left).
For a given state $s_t$, the status point is initially located at the root ($node_1$) and moves to one of its child nodes ($node_3$) according to the probability distribution given by the policy network corresponding to the root ($node_1$).
And the status point keeps moving until reaching a leaf node and the corresponding item ($item_8$ in Figure \ref{fig:TPGR}) is recommended to the user.

\begin{algorithm}[!t]
	\caption{Learning TPGR}
	\label{alg:TPGR}
	\begin{algorithmic}[1]
		\REQUIRE
		episode length $n$, tree depth $d$, discount factor $\gamma$, learning rate $\eta$, reward function $\mathcal{R}$, item set $A$ with representation vectors
		\ENSURE
		model parameters $\theta$
		\STATE{$c = ceil(|A|^{\frac{1}{d}})$}
		\STATE{construct a balanced clustering tree $T$ with the number of child nodes set to $c$}
		\STATE{$\mathcal{I}=\frac{c^d-1}{c-1}$}
		\FOR{$j=1$ to $\mathcal{I}$}
		\STATE{initialize $\theta_j\leftarrow$ random values}
		\ENDFOR
		\STATE{$\theta=(\theta_1, \theta_2, ... , \theta_\mathcal{I})$}
		\REPEAT
		\STATE{$\Delta\theta = 0$}
		\STATE{$(s_1, p_1, r_1, ... , s_n, p_n, r_n)\leftarrow$ SamplingEpisode($\theta$, $n$, $c$, $d$, $T$, $\mathcal{R}$) (see Algorithm \ref{alg:sample2})}
		\FOR{$t=1$ to $n$}
		\STATE{map $p_t$ to an item $a_t$ w.r.t. $T$ and record the trajectory nodes' indexes $(i_1, i_2, ... , i_d)$}
		\STATE{$\hat{Q}^{\pi_\theta}(s_t, a_t)=\sum_{i=t}^n \gamma^{i-t}r_i$}
		\STATE{$\pi_\theta(a_t|s_t) = \prod_{d'=1}^{d}\pi_{\theta_{i_{d'}}}(p_{td'}|s_t)$}
		\STATE{$\Delta\theta = \Delta\theta + \nabla_\theta \log \pi_\theta(a_t|s_t)\hat{Q}^{\pi_\theta}(s_t, a_t)$}
		\STATE{$\theta=\theta+\eta\Delta\theta$}
		\ENDFOR		
		\UNTIL{converge}
		\RETURN{$\theta$}
	\end{algorithmic}
\end{algorithm}

We use the REINFORCE algorithm \cite{williams1992simple} to train the model while other policy gradient algorithms can be utilized analogously. The objective is to maximize the expected discounted cumulative rewards, i.e.,
\begin{align}
J(\pi_\theta) = \mathbb{E}_{\pi_\theta}\Big[ \sum_{i=1}^{n} \gamma^{i-1} r_i \Big],
\end{align}
\noindent and one of its approximate gradient with respect to the parameters is:
\begin{align}
\nabla_\theta J(\pi_\theta) \approx \mathbb{E}_{\pi_\theta}[\nabla_\theta \log \pi_\theta(a|s) Q^{\pi_\theta}(s, a)],
\end{align}
where $\pi_\theta(a | s)$ is the probability of taking the action $a$ at the state $s$, and $Q^{\pi_\theta}(s, a)$ denotes the expected discounted cumulative rewards starting with $s$ and $a$, which can be estimated empirically by sampling trajectories following the policy $\pi_\theta$. 

An algorithmic description of the training procedure is given in Algorithm~\ref{alg:TPGR} where $\mathcal{I}$ denotes the number of non-leaf nodes of the tree.
When sampling an episode for TPGR (as shown in Algorithm~\ref{alg:sample2}), $p_t$ denotes the path from the root to a leaf at timestep $t$, which consists of $d$ choices, and each choice is represented as an integer between $1$ and $c$ denoting the corresponding child node to move.
Making the consecutive choices corresponding to $p_t$ from the root, we traverse the nodes along $p_t$ and finally reach a leaf node. As such, a path $p_t$ is mapped to a recommended item $a_t$, thus the probability of choosing $a_t$ given state $s_t$ is the product of the probability of making each choice (to reach $a_t$) along $p_t$. 

\subsubsection{Time and Space Complexity Analysis}
\begin{algorithm}[!t]
	\caption{Sampling Episode for TPGR}
	\label{alg:sample2}	
	\begin{algorithmic}[1]
		\REQUIRE
		parameters $\theta$, episode length $n$, maximum child number $c$, tree depth $d$, balanced clustering tree $T$, reward function $\mathcal{R}$
		\ENSURE
		an episode $E$
		\STATE{Initialize $s_1 \leftarrow [0]$}
		\FOR{$t=1$ to $n$}
		\STATE{$node\_index = 1$}
		\FOR{$d'=1$ to $d$}
		\STATE{sample $c_{d'}\sim\pi_{\theta_{node\_index}}(s_t)$}
		\STATE{$node\_index=(node\_index-1)\times c + c_{d'} + 1$}
		\ENDFOR
		\STATE{$p_t = (c_1, c_2, ... ,c_d)$}
		\STATE{map $p_t$ to an item $a_t$ w.r.t. $T$}
		\STATE{$r_t = \mathcal{R}(s_t, a_t)$}
		\IF{$t<n$}
		\STATE{calculate $s_{t+1}$ as described in Figure \ref{fig:state representation}}
		\ENDIF
		\ENDFOR
		\RETURN{$E=(s_1, p_1, r_1, ... , s_n, p_n, r_n)$}
	\end{algorithmic}
\end{algorithm}

Empirically, the value of the tree depth $d$ is set to a small constant (typically set to 2 in our experiments).
Thus, both the time (for making a decision) and the space complexity of each policy network is $\mathcal{O}(c)$ (see more details in the appendices).

Considering the time spent on sampling an action given a specific state in Algorithm \ref{alg:sample2}, the TPGR makes $d$ choices, each of which is based on a policy network with at most $c$ output units.
Therefore, the time complexity of sampling one item in the TPGR is $\mathcal{O}(d \times c) \simeq \mathcal{O}(d \times |A|^\frac{1}{d})$. 
Compared to the normal RL-based methods whose time complexity of sampling an action is $\mathcal{O}(|A|)$, our proposed TPGR can significantly reduce the time complexity.

The space complexity of each policy network is $\mathcal{O}(c)$ and the number of non-leaf nodes (i.e., the number of policy networks) of the constructed clustering tree is:
\begin{align}
\mathcal{I} = 1 + c + c^2 + \dots + c^{d-1} = \frac{c^d-1}{c-1}.
\end{align}

Therefore, the space complexity of the TPGR is $\mathcal{O}(\mathcal{I}\times c) \simeq \mathcal{O}(\frac{c^d-1}{c-1}\times c) \simeq \mathcal{O} (c^d) \simeq \mathcal{O} (|A|)$, which is the same as that of normal RL-based methods.

\subsection{State Representation}
\begin{figure}[!htb]
	\centering
	\includegraphics[width=0.86\columnwidth]{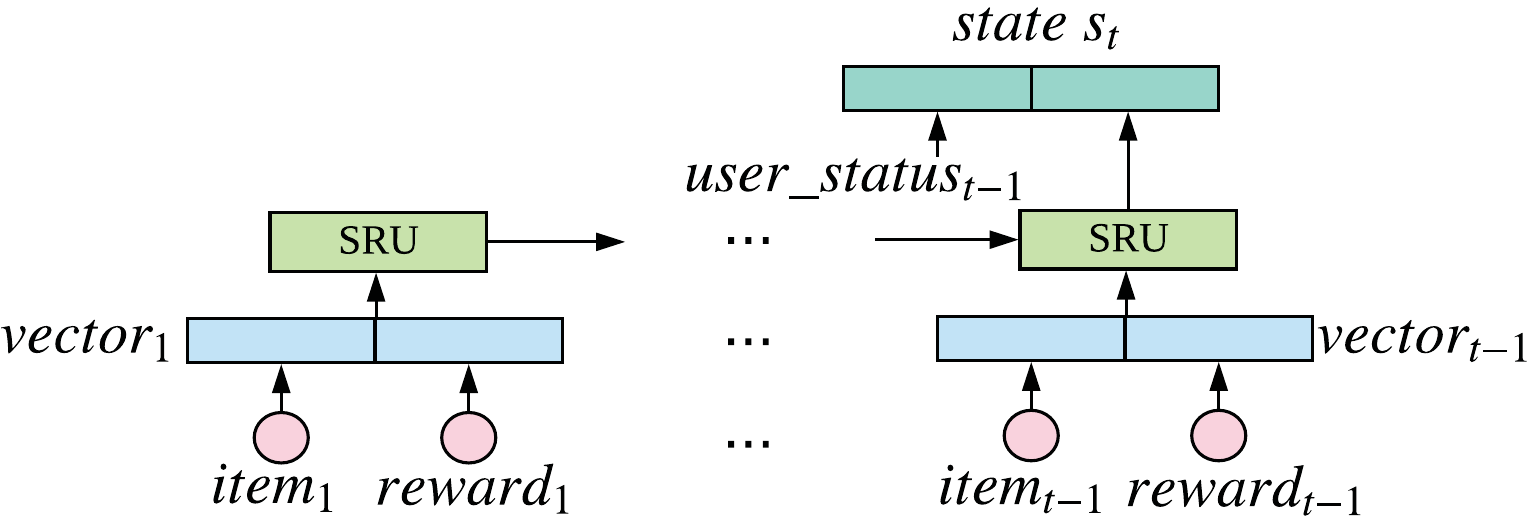}	\caption{State representation.}
	\label{fig:state representation}
\end{figure}

In this section, we present the state representation scheme adopted in this work, whose details are shown in Figure~\ref{fig:state representation}.

In Figure~\ref{fig:state representation}, we assume that the recommender system is performing the $t$-th recommendation.
The input is a sequence of recommended item IDs and the corresponding rewards (user's feedbacks) before timestep $t$.
Each item ID is mapped to an embedding vector which can be learned together with the policy networks in an end-to-end manner, or can be pre-trained by some supervised learning models such as matrix factorization and is fixed while training.
Each reward is mapped to a one-hot vector with a simple reward mapping function (see more details in the appendices).

For encoding the historical interactions, we adopt a simple recurrent unit (SRU) \cite{lei2017training}, an RNN model that is fast to train, to learn the hidden representation.
Besides, to further integrate more feedback information, we construct a vector, denoted as $user\_status_{t-1}$ in Figure \ref{fig:state representation}, containing some statistic information such as the number of positive rewards, negative rewards, consecutive positive and negative rewards before timestep $t$, which is then concatenated with the hidden vector generated by the SRU to gain the state representation at timestep $t$.

\section{Experiments and Results}\label{sec:exp}

\subsection{Datasets}
We adopt the following two datasets in our experiments.
\begin{itemize}
	\item \textbf{MovieLens (10M).\footnote{http://files.grouplens.org/datasets/movielens/ml-10m.zip}} A dataset consists of 10 million ratings from users to movies in MovieLens website.
	\item \textbf{Netflix.\footnote{https://www.kaggle.com/netflix-inc/netflix-prize-data}} A dataset contains 100 million ratings from Netflix's competition to improve their recommender systems. 
\end{itemize}
Detailed statistic information, including the number of users, items and ratings, of these datasets is given in Table \ref{tab:info}.

\begin{table}[h]
	\centering
	\caption{Statistic information of the datasets.}
	\label{tab:info}
	\resizebox{1\columnwidth}{!}{  
		\begin{tabular}{c|ccccc}
			\toprule
			Dataset &\#users &\#items &\tabincell{c}{total \\ \#ratings} &\tabincell{c}{\#ratings \\ per user} &\tabincell{c}{\#ratings \\per item} \\
			\midrule
			MovieLens &69,878 &10,677 &10,000,054 &143 &936 \\
			Netflix &480,189 &17,770 &100,498,277 &209 &5,655 \\
			\bottomrule
		\end{tabular}
	}
\end{table}

\subsection{Data Analysis}\label{sec:da}
To demonstrate the existence of hidden sequential patterns in the recommendation process, we empirically analyze the aforementioned two datasets where each rating is attached with a timestamp.
Each dataset comprises numerous user sessions and each session contains the ratings from one specific user to various items along timestamps.

Without loss of generality, we regard the ratings higher than 3  as positive ratings (noticed that the highest rating is 5) and the others as negative ratings.
For a rating with at most $b$ consecutive positive (negative) ratings before it, we define its consecutive positive (negative) count as $b$. As such, each rating can be associated with a specific consecutive positive (negative) count and we can calculate the average rating for ratings with the same consecutive positive (negative) count.

\begin{figure}[!htb]
	\centering
	\includegraphics[width=0.49\columnwidth]{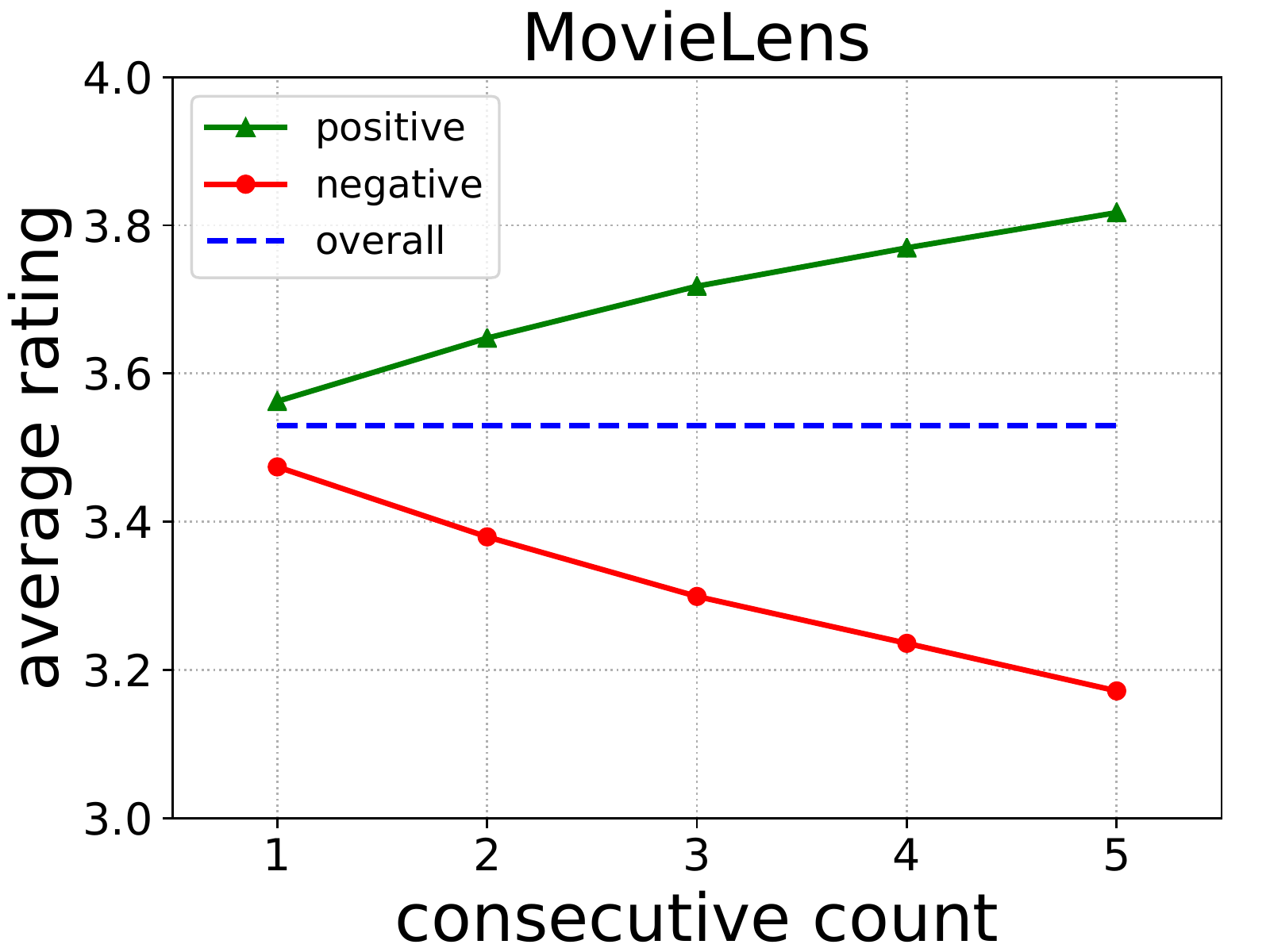}
	\includegraphics[width=0.49\columnwidth]{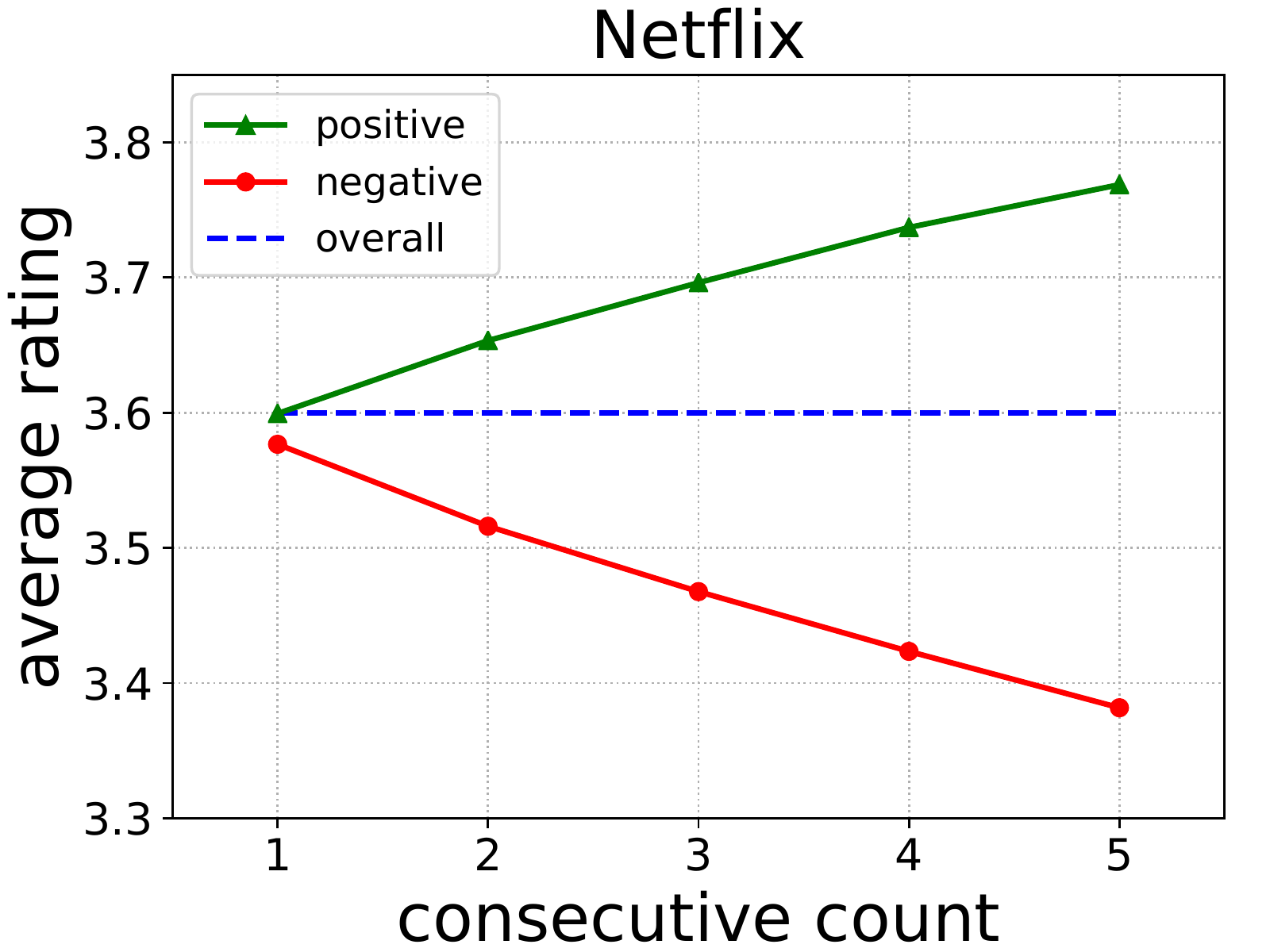}	\caption{Average ratings for different consecutive counts.}
	\label{fig:rating}
\end{figure}

We present the corresponding average ratings w.r.t. the consecutive positive (negative) counts in Figure~\ref{fig:rating}, where we can clearly observe the sequential patterns in the user's rating behavior: a user tends to give a linearly higher rating for an item with larger consecutive positive count (green line) and vice versa (red line). 
The reason may be that the more satisfying (disappointing) items a user has consumed before, the more pleasure (displeasure) she gains and as a result, she tends to give a higher (lower) rating to the current item.

\begin{table*}[!t]
	\centering
	\caption{Overall interactive recommendation performance (* indicates that p-value is less than $10^{-6}$ for significance test).}\label{tab:main}
	\resizebox{2.1\columnwidth}{!}{
		\begin{tabular}{c|c|c|c|c|c|c|c|c|c|c|c|c|c}
			\toprule
			\multirow{2}{*}{Dataset} &\multirow{2}{*}{Method} &\multicolumn{4}{c|}{$\alpha=0.0$} &\multicolumn{4}{c|}{$\alpha=0.1$} &\multicolumn{4}{c}{$\alpha=0.2$} \\ \cmidrule{3-14}
			&&Reward&Precision@$32$&Recall@$32$&F1@$32$&Reward&Precision@$32$&Recall@$32$&F1@$32$&Reward&Precision@$32$&Recall@$32$&F1@$32$ \\ 
			\midrule
			\multirow{8}{*}{MovieLens} 
			&Popularity &0.0315 &0.0405 &0.0264 &0.0257 &0.0349 &0.0405 &0.0264 &0.0257 &0.0383 &0.0405 &0.0264
			&0.0257 \\
			&GreedySVD &0.0561 &0.0756 &0.0529 &0.0498 &0.0655 &0.0759 &0.0532 &0.0501 &0.0751 &0.0760 &0.0532 &0.0502 \\
			&LinearUCB &0.0680 &0.0920 &0.0627 &0.0597 &0.0798 &0.0919 &0.0627 &0.0597 &0.0917 &0.0919 &0.0627 &0.0598 \\
			&HLinearUCB &0.0847 &0.1160 &0.0759 &0.0734 &0.1023 &0.1162 &0.0759 &0.0735 &0.1196 &0.1165 &0.0761 &0.0737 \\
			&DDPG-KNN($k=1$) &0.0116 &0.0234 &0.0082 &0.0098 &0.0143 &0.0240 &0.0086 &0.0102 &0.0159 &0.0239 &0.0086 &0.0102 \\
			&DDPG-KNN($k=0.1N$) &0.1053 &0.1589 &0.0823 &0.0861 &0.1504 &0.1754 &0.0918 &0.0964 &0.1850 &0.1780 &0.0922 &0.0975 \\			
			&DDPG-KNN($k=N$) &0.1764 &0.2605 &0.1615 &0.1562 &0.2379 &0.2548 &0.1529 &0.1504 &0.3029 &0.2542 &0.1437 &0.1477 \\
			&DDPG-R &0.0898 &0.1396 &0.0647 &0.0714 &0.1284	&0.1639	&0.0798	&0.0862 &0.1414	&0.1418	&0.0656	&0.0724 \\
			&DQN-R &0.1610 &0.2309 &0.1304 &0.1326 &0.2243 &0.2429 &0.1466 &0.1450 &0.2490 &0.2140 &0.1170 &0.1204 \\
			&TPGR &\textbf{0.1861}* &\textbf{0.2729}* &\textbf{0.1698}* &\textbf{0.1666}* &\textbf{0.2472}* &\textbf{0.2726}* &\textbf{0.1697}* &\textbf{0.1665}* &\textbf{0.3101} &\textbf{0.2729}* &\textbf{0.1702}* &\textbf{0.1667}* \\
			\midrule
			\multirow{8}{*}{Netflix} 
			&Popularity &0.0000 &0.0002 &0.0001 &0.0001 &0.0000 &0.0002 &0.0001 &0.0001 &0.0000 &0.0002 &0.0001 &0.0001 \\
			&GreedySVD &0.0255 &0.0320 &0.0113 &0.0132 &0.0289 &0.0327 &0.0115 &0.0135 &0.0310 &0.0315 &0.0113 &0.0132 \\
			&LinearUCB &0.0557 &0.0682 &0.0212 &0.0263 &0.0652 &0.0681 &0.0212 &0.0263 &0.0744 &0.0679 &0.0211 &0.0262 \\
			&HLinearUCB &0.0800 &0.1005 &0.0314 &0.0387 &0.0947 &0.0999 &0.0312 &0.0385 &0.1077 &0.0995 &0.0310 &0.0382 \\
			&DDPG-KNN($k=1$) &0.0195 &0.0291 &0.0092 &0.0106 &0.0252 &0.0328 &0.0096 &0.0113 &0.0272 &0.0314 &0.0094 &0.0111 \\
			&DDPG-KNN($k=0.1N$) &0.1127 &0.1546 &0.0452 &0.0561 &0.1581 &0.1713 &0.0546 &0.0653 &0.1848 &0.1676 &0.0517 &0.0632 \\			
			&DDPG-KNN($k=N$) &0.1355 &0.1750 &0.0447 &0.0598 &0.1770 &0.1745 &0.0521 &0.0646 &0.2519 &0.1987 &0.0584 &0.0739 \\
			&DDPG-R &0.1008 &0.1300 &0.0343 &0.0441 &0.1127 &0.1229 &0.0327 &0.0420 &0.1412 &0.1263 &0.0351 &0.0445 \\
			&DQN-R &0.1531 &0.2029 &0.0731 &0.0824 &0.2044 &0.1976 &0.0656 &0.0757 &0.2447 &0.1927 &0.0526 &0.0677 \\
			&TPGR &\textbf{0.1881}* &\textbf{0.2511}* &\textbf{0.0936}* &\textbf{0.1045}* &\textbf{0.2544}* &\textbf{0.2516}* &\textbf{0.0921}* &\textbf{0.1037}* &\textbf{0.3171}* &\textbf{0.2483}* &\textbf{0.0866}* &\textbf{0.1003}* \\ 
			\bottomrule
		\end{tabular}
	}
\end{table*}

\subsection{Environment Simulator and Reward Function}\label{sec:reward}
To train and test RL-based recommendation algorithms, a straightforward way is to conduct online experiments where the recommender system can directly interact with real users, which, however, could be too expensive and commercially risky for the platform \cite{zhang2016collective}.  Thus, in this paper, we focus on evaluating our proposed model on public available offline datasets by building up an environment simulator to mimic online environments. 

Specifically, we normalize the ratings of a dataset into range $[-1, 1]$ and use the normalized value as the empirical reward of the corresponding recommendation.
To take the sequential patterns into account, we combine a sequential reward with the empirical reward to construct the final reward function.
Within each episode, the environment simulator randomly samples a user $i$ and the recommender system starts to interact with the sampled user $i$ until the end of the episode, and the reward of recommending item $j$ to user $i$, denoted as action $a$, at state $s$ is given as:
\begin{align}
\mathcal{R}(s, a) = r_{ij} + \alpha \times (c_p - c_n),
\end{align}
where $r_{ij}$ is the corresponding normalized rating and is set to $0$ if user $i$ does not rate item $j$ in the dataset,  $c_p$ and $c_n$ denote the consecutive positive and negative counts respectively; $\alpha$ is a non-negative parameter to control the trade-off between the empirical reward and the sequential reward.

\subsection{Main Experiments}

\subsubsection{Compared Methods}
We compare our TPGR model with 7 methods in our experiments where {Popularity} and {GreedySVD} are conventional recommendation methods; {LinearUCB} and {HLinearUCB} are MAB-based methods; {DDPG-KNN}, {DDPG-R} and {DQN-R} are RL-based methods.
\begin{itemize}
	\item \textbf{Popularity} recommends the most popular item (i.e., the item with highest average rating) from current available items to the user at each timestep.
	\item \textbf{GreedySVD} trains the singular value decomposition (SVD) model after each interaction and picks the item with highest rating predicted by the SVD model.
	\item \textbf{LinearUCB} is a contextual-bandit recommendation approach \cite{li2010contextual} which adopts a linear model to estimate the upper confidence bound (UCB) for each arm.
	\item \textbf{HLinearUCB} is also a contextual-bandit recommendation approach \cite{wang2016learning} which  learns extra hidden features for each arm to model the reward.
	\item \textbf{DDPG-KNN} denotes the method \cite{dulac2015deep} addressing the large discrete action space problem by combining DDPG with an approximate KNN method.
	\item \textbf{DDPG-R} denotes the DDPG-based recommendation method \cite{zhao2018deep} which learns a ranking vector and picks the item with highest ranking score.
	\item \textbf{DQN-R} denotes the DQN-based recommendation method \cite{zheng2018drn} which utilizes a DQN to estimate Q-value for each action given the current state.
\end{itemize}
\subsubsection{Experiment Details}
For each dataset, the users are randomly divided into two parts where 80\% of the users are used for training while the other 20\% are used for test. 
In our experiments, the length of an episode is set to 32 and the trade-off factor $\alpha$ in the reward function is set to $0.0$, $0.1$ and $0.2$ respectively for both datasets.
In each episode, once an item is recommended, it is removed from the set of available items, thus no repeated items occur in an episode. 

For DDPG-KNN, larger $k$ (i.e., the number of nearest neighbors) leads to better performance but poorer efficiency and vice versa \cite{dulac2015deep}.
For fair comparison, we consider three cases with the value of $k$ set to 1, 0.1$N$ and $N$ ($N$ denotes the number of items) respectively.

For TPGR, we set the clustering tree depth $d$ to 2 and apply the PCA-based clustering algorithm with rating-based item representation when constructing the balanced tree since they give the best empirical results as shown in the following section. The implementation code\footnote{https://github.com/chenhaokun/TPGR} of the TPGR is available online.

All other hyper-parameters of all the models are carefully chosen by grid search.

\subsubsection{Evaluation Metrics}
As the target of RL-based methods is to gain the optimal long-run rewards, we use the average reward over each recommendation for each user in test set as one evaluation metric. Furthermore, we adopt Precision@$k$, Recall@$k$ and F1@$k$ \cite{Herlocker2004Evaluating} as our evaluation metrics. Specifically, we set the value of $k$ as 32, which is the same as the episode length. For each user, all the items with a rating higher than 3.0 are regarded as the relevant items while the others are regarded as the irrelevant ones.

\subsubsection{Results and Analysis}
In our experiments, all the models are evaluated in term of the four metrics including average reward over each recommendation, Precision@$32$, Recall@$32$, and F1@$32$.
The summarized results are presented in Table \ref{tab:main} with respect to the two datasets and three different settings of trade-off factor $\alpha$ in the reward function.

From Table \ref{tab:main}, we observe that our proposed TPGR outperforms all the compared methods in all settings with p-values less than $10^{-6}$ (indicated by a * mark in Table \ref{tab:main}) for significance test \cite{Ruxton2006The} in most cases, which demonstrates the performance superiority of the TPGR.

When comparing the RL-based methods with the conventional and the MAB-based methods, it is not surprising to find that the RL-based models provide superior performances in most cases, as they have the ability of long-run planning and dynamic adaptation which is lacking in other methods.
Among all the RL-based methods, our proposed TPGR achieves the best performance, which can be explained by two reasons.
First, the hierarchical clustering over items incorporates additional item similarity information into our model, e.g., similar items tend to be clustered into one subtree of the clustering tree.
Second, different from normal RL-based methods which utilize one complicated neural network to make decisions, we propose to conduct a tree-structured decomposition and adopt a certain number of policy networks with much simpler architectures, which may ease the training process and lead to better performance.

Besides, as the value of trade-off factor $\alpha$ increases, we observe that the improvement of TPGR over HLinearUCB (i.e., the best non-RL-based method in our experiments) in terms of average reward becomes more significant, which demonstrates that the TPGR do have the capacity of capturing sequential patterns to maximize long-run rewards.

\subsubsection{Time Comparison}
In this section, we compare the efficiency (in term of the consumed time for training and decision making stages) of RL-based models on the two datasets.

To make the time comparison fair, we remove the limitation of no repeated items in an episode to avoid involving the masking mechanism as the efficiency of the different implementations of the masking mechanism is highly different.
Besides, all the models adopt the neural networks with the same architecture which consists of three fully-connected layers with the numbers of hidden units set to 32 and 16 respectively, and the experiments are conducted on the same machine with 4-core 8-thread CPU (i7-4790k, 4.0GHz) and 32GB RAM.
We record the consumed time for one training step (i.e., sampling 1 thousand episodes and updating the model with those episodes) and the consumed time for making 1 million decisions for each model.

As shown in Table~\ref{tab:time}, TPGR consumes much less time for both the training and the decision making stages compared to DQN-R and DDPG-R.
DDPG-KNN with $k$ set to 1 gains high efficiency, which, however, is meaningless because it achieves very poor recommendation performance as shown in Table~\ref{tab:main}.
In another case where $k$ is set to $N$, DDPG-KNN suffers from high time complexity which makes it even much slower than DQN-R and DDPG-R.
Thus, DDPG-KNN can not achieve high effectiveness and high efficiency at the same time.
Compared to the case that DDPG-KNN makes a trade-off between effectiveness and efficiency, i.e., setting $k$ as $0.1N$, our proposed TPGR achieves significant improvement in term of both effectiveness and efficiency.
\begin{table}[!t]
	\centering
	\caption{Time comparison for training and decision making.}\label{tab:time}
	\resizebox{1\columnwidth}{!}{
		\begin{tabular}{c|c|c|c|c}
			\toprule
			\multirow{2}{*}{Method} &\multicolumn{2}{c|}{Seconds per training step} &\multicolumn{2}{c}{Seconds per $10^6$ decisions}  \\
			\cmidrule{2-5}
			&MovieLens &Netflix &MovieLens &Netflix \\
			\midrule
			DQN-R &13.1 &15.3 &19.6 &34.6 \\
			DDPG-R &44.6 &58.6 &29.4 &49.6 \\
			DDPG-KNN($k=1$) &1.3 &1.3 &1.8 &1.8 \\
			DDPG-KNN($k=0.1N$) &24.2 &40.3 &200.4 &313.0 \\
			DDPG-KNN($k=N$) &248.4 &323.9 &1,875.0 &3,073.2 \\
			TPGR &3.0 &3.1 &3.4 &3.9 \\
			\bottomrule
		\end{tabular}
	}
\end{table}

\subsection{Influence of Clustering Approach}\label{sec:clustering}
\begin{figure}[!htb]
	\centering
	\includegraphics[width=0.49\columnwidth]{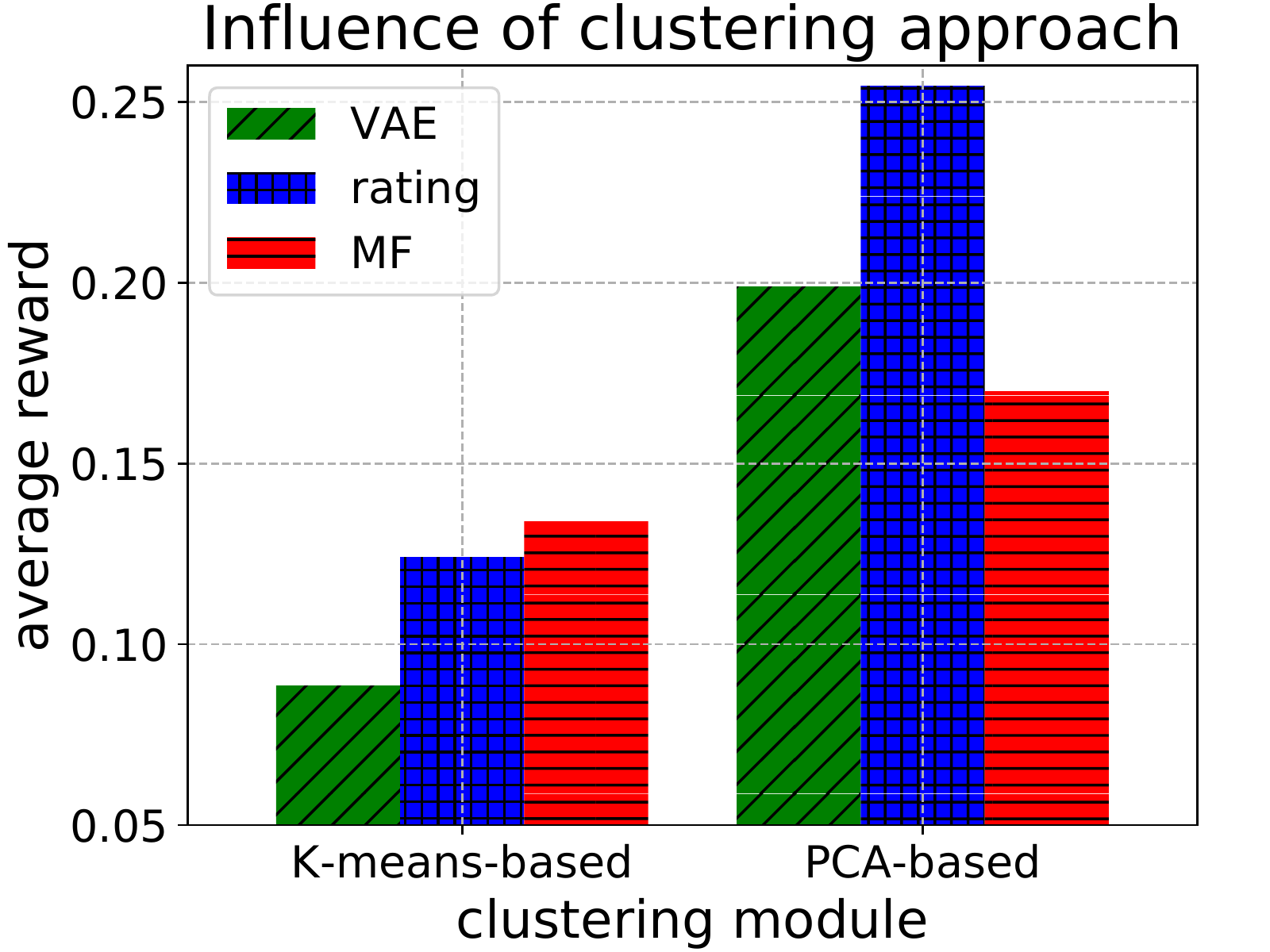}
	\includegraphics[width=0.49\columnwidth]{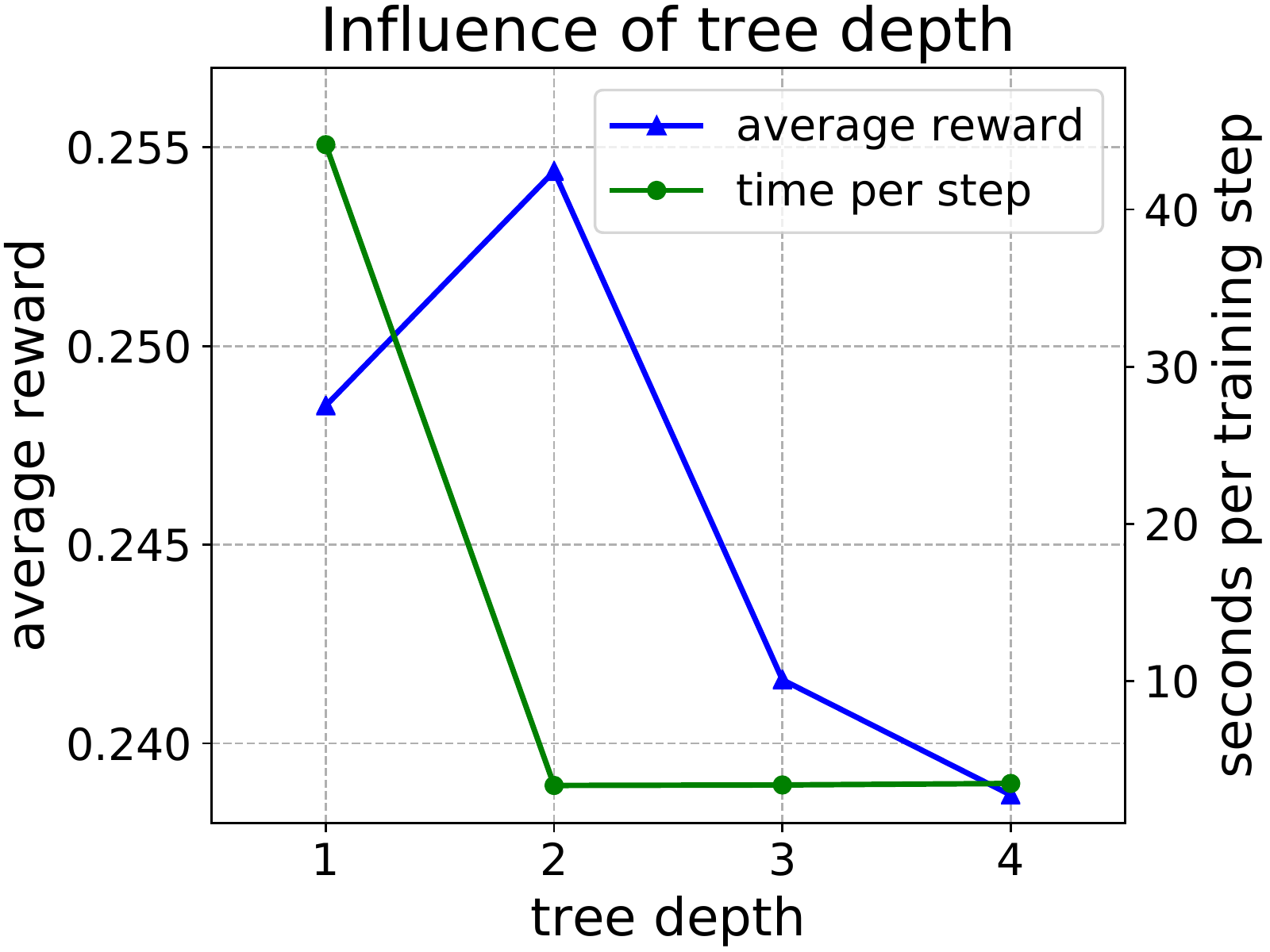}
	\caption{Influence of clustering approach and tree depth.}
	\label{fig:influence}
\end{figure}

Since the architecture of the TPGR is based on the balanced hierarchical clustering tree, it is essential to choose a suitable clustering approach. 
In the previous section, we introduce two clustering modules, K-means-based and PCA-based modules, and three methods to represent an item, namely rating-based, MF-based and VAE-based methods.
As such, there are six combinations to conduct balanced hierarchical clustering.
With $\alpha$ set to $0.1$, we evaluate the above six approaches in term of average reward on Netflix dataset.
The results are shown in Figure~\ref{fig:influence} (left).

As shown in Figure \ref{fig:influence} (left), applying PCA-based clustering module with rating-based item representation achieves the best performance.
Two reasons may account for this result.
First, the rating-based representation retains all the interaction information between the users and the items, while both the VAE-based and the MF-based representations are low-dimensional, which retain less information than rating-based representation after dimension reduction.
Therefore, using rating-based representation may lead to better clustering. 
Second, as the number of clusters $c$ (i.e., child nodes number of non-leaf nodes) is large (134 for Netflix dataset with the tree depth set to 2), the quality of the clustering tree derived from K-means-based method would be sensitive to the choices of the initialization of center points and the distance function, etc., which may lead to worse performance than more robust methods such as PCA-based method, as observed in our experiments. 

\subsection{Influence of Tree Depth}\label{sec:depth}
To show how the tree depth influences the performance as well as the training time of the TPGR, we vary the tree depth from 1 to 4 and record the corresponding results.

As shown in Figure \ref{fig:influence} (right), the green curve shows the consumed time per training step with respect to different tree depths, where each training step consists of sampling 1 thousand episodes and updating the model with those episodes.
It should be noticed that the model with tree depth set to $1$ is actually without a tree structure but with only one policy network taking a state as input and giving the policy possibility distribution over all items.
Thus, the tree-structured models (i.e., models with tree depth set to 2, 3 and 4) do significantly improve the efficiency.
The blue curve in Figure \ref{fig:influence} (right) presents the performance of the TPGR over different tree depths, from which we can see that the model with tree depth set to 2 achieves the best performance while other tree depths lead to a slight discount on performance.
Therefore, setting the depth of the clustering tree to 2 is a good starting point to explore suitable tree depth when using the TPGR, which can significantly reduce the time complexity and provide great or even the best performance.

\section{Conclusion}\label{sec:conclusion}
In this paper, we propose a Tree-structured Policy Gradient Recommendation (TPGR) framework to conduct large-scale interactive recommendation. TPGR performs balanced hierarchical clustering over the discrete action space
to reduce the time complexity of RL-based recommendation methods, which is crucial for scenarios with a large number of items.
Besides, it explicitly models the long-run rewards and captures the sequential patterns so as to achieve higher rewards in the long run.
Thus, TPGR has the capacity of achieving high efficiency and high effectiveness at the same time.
Extensive experiments over a carefully-designed simulator based on two public datasets demonstrate that the proposed TPGR, compared to the state-of-the-art models, can lead to better performance with higher efficiency. 
For future work, we plan to deploy TPGR onto an online commercial recommender system.
We also plan to explore more clustering tree construction schemes based on the current recommendation policy, which is also a fundamental problem for large-scale discrete action clustering in reinforcement learning. 

\section{Acknowledgements}
The work is sponsored by Huawei Innovation Research Program. The corresponding author Weinan Zhang thanks the support of National Natural Science Foundation of China (61632017, 61702327, 61772333), Shanghai Sailing Program (17YF1428200).

\section{Appendices}
\subsection{Clustering Modules}
We introduce two balanced clustering modules in this paper, namely, K-means-based and PCA-based modules, whose algorithmic details are shown in Algorithm~\ref{alg:k-means} and Algorithm~\ref{alg:pca} respectively.
\begin{algorithm}[]
	\caption{K-means-based Balanced Clustering}
	\label{alg:k-means}	
	\begin{algorithmic}[1]
		\REQUIRE
		a group of vectors $v_1, v_2, ... ,v_m$ and the number of clusters $c$
		\ENSURE
		clustering result
		\FOR{$j=1$ to $c$}
		\STATE{initialize the $j^{th}$ cluster $\leftarrow \emptyset$ }
		\ENDFOR
		\IF{$m\leq c$}
		\FOR{$j=1$ to $m$}
		\STATE{assign $v_j$ to the $j^{th}$ cluster}
		\ENDFOR
		\RETURN{first $m$ clusters}
		\ENDIF
		\STATE{use the normal k-means algorithm to find $c$ centroids: $p_1, p_2, ... , p_{c}$}
		\STATE{mark all input vectors as unassigned}
		\STATE{$i = 1$}
		\WHILE{not all vectors are marked as assigned}
		\STATE{find the vector $v'$ among unassigned vectors which is with the shortest Euclid distance to $p_i$}
		\STATE{assign $v'$ to the $i^{th}$ cluster}
		\STATE{mark $v'$ as assigned}
		\STATE{$i = i\enspace mod$ $c + 1$}
		\ENDWHILE
		\RETURN{all $c$ clusters}
	\end{algorithmic}
\end{algorithm}
		
\begin{algorithm}[!t]
	\caption{PCA-based Balanced Clustering}
	\label{alg:pca}	
	\begin{algorithmic}[1]
		\REQUIRE
		a group of vectors $v_1, v_2, ... ,v_m$ and the number of clusters $c$
		\ENSURE
		clustering result
		\FOR{$j=1$ to $c$}
		\STATE{initialize the $j^{th}$ cluster $\leftarrow \emptyset$ }
		\ENDFOR
		\IF{$m\leq c$}
		\FOR{$j=1$ to $m$}
		\STATE{assign $v_j$ to the $j^{th}$ cluster}
		\ENDFOR
		\RETURN{first $m$ clusters}
		\ENDIF
		\STATE{use PCA to find the principal component $u$ with the largest possible variance}
		\STATE{sort the input vectors according to the value of projections on $u$ and gain $v_{i_1}, v_{i_2}, ... ,v_{i_m}$}
		\STATE{$thredhold= (m-1) mod$ $c + 1$}
		\STATE{$max\_length = ceil(m/c)$}
		\FOR{$j=1$ to $thredhold$}
		\STATE{$start = (j-1)\times max\_length+1$}
		\STATE{assign $v_{start}, v_{start+1}, ... , v_{start+max\_length-1}$ to the $j^{th}$ cluster}
		\ENDFOR
		\FOR{$j=thredhold+1$ to $c$}
		\STATE{$start = thredhold\times max\_length + (j-1-thredhold)\times (max\_length-1) + 1$}
		\STATE{assign $v_{start}, v_{start+1}, ... , v_{start+max\_length-2}$ to the $j^{th}$ cluster}
		\ENDFOR
		\RETURN{all $c$ clusters}
	\end{algorithmic}
\end{algorithm}
		
\subsection{Time and Space Complexity for Each Policy Network of TPGR}
As the value of the tree depth $d$ is empirically set to a small constant (typically set to 2 in our experiments) and $c$ equals to $ceil(|A|^{\frac{1}{d}})$, we have: 
\begin{align}
\label{c+m}
\mathcal{O}(c+m) \simeq \mathcal{O}(|A|^{\frac{1}{d}}+m) \simeq \mathcal{O}(|A|^{\frac{1}{d}}) \simeq \mathcal{O}(c)
\end{align}
and
\begin{align}
\label{cm}
\mathcal{O}(m\times c) \simeq \mathcal{O}(m\times |A|^{\frac{1}{d}}) \simeq \mathcal{O}(|A|^{\frac{1}{d}}) \simeq \mathcal{O}(c)
\end{align}
where $m$ is a constant.

As described in the paper, each policy network is implemented as a fully-connected neural network.
Thus, the time complexity of making a decision for each policy network is $\mathcal{O}(a + b\times c)$, where $a$ is a constant indicating the time consuming before the output layer while $b$ is also a constant indicating the number of hidden units of the hidden layer before the output layer.
According to Eq.~\ref{c+m} and Eq.~\ref{cm}, we have $\mathcal{O}(a + b\times c) \simeq \mathcal{O}(c)$.

A similar analysis can be applied to derive the space complexity for each policy network. Assuming that the space occupation for each policy network except the parameters of the output layer is $a'$ and the number of hidden units of the hidden layer before the output layer is $b'$, we can derive that the space complexity for each policy network is $\mathcal{O}(a' + b'\times c) \simeq \mathcal{O}(c)$.

Thus, both the time (for making a decision) and the space complexity of each policy network is linear to the size of its output units, i.e., $\mathcal{O}(c)$.

\subsection{Reward Mapping Function}
Assuming that the range of reward values is $(a, b]$ and the desired dimension of the one-hot vector is $l$, we define the reward mapping function as:
\begin{align*}
onehot\_mapping(r) = onehot\Big(l - floor\big(\frac{l \times (b - r)}{b - a}\big), l\Big)
\end{align*}
where $floor(x)$ returns the largest integer no greater than $x$ and $one\_hot(i, l)$ returns an $l$-dimensional vector where the value of the $i$-th element is 1 while the others are set to 0.

\bibliography{tpgr}
\bibliographystyle{aaai}
\end{document}